\documentclass{article}

\usepackage{arxiv}

\usepackage[utf8]{inputenc} 
\usepackage[T1]{fontenc}    
\usepackage{hyperref}       
\usepackage{url}            
\usepackage{booktabs}       
\usepackage{amsfonts}       
\usepackage{nicefrac}       
\usepackage{microtype}      
\usepackage{lipsum}
\usepackage{multirow}
\usepackage[table,xcdraw]{xcolor}
\usepackage{booktabs}
\usepackage[pdftex]{graphicx}
\usepackage{xcolor}

\title{Grid Search Hyperparameter Benchmarking of BERT, ALBERT, and LongFormer on DuoRC}

\author{
Alex John Quijano$^{a}$ \hspace{10px} Sam Nguyen$^{b}$ \hspace{10px} Juanita Ordonez$^{b}$\\
  $^{a}$University of California Merced \hspace{10px}\\
  $^{b}$Lawrence Livermore National Laboratory\\
}

\begin{document}
\maketitle

\begin{abstract}
The purpose of this project is to evaluate three language models named BERT, ALBERT, and LongFormer on the Question Answering dataset called DuoRC. The language model task has two inputs, a question, and a context. The context is a paragraph or an entire document while the output is the answer based on the context. The goal is to perform grid search hyperparameter fine-tuning using DuoRC. Pretrained weights of the models are taken from the Huggingface library. Different sets of hyperparameters are used to fine-tune the models using two versions of DuoRC which are the SelfRC and the ParaphraseRC. The results show that the ALBERT (pretrained using the SQuAD1 dataset) has an F1 score of 76.4 and an accuracy score of 68.52 after fine-tuning on the SelfRC dataset. The Longformer model (pretrained using the SQuAD and SelfRC datasets) has an F1 score of 52.58 and an accuracy score of 46.60 after fine-tuning on the ParaphraseRC dataset. The current results outperformed the results from the previous model by DuoRC.
\end{abstract}



\section{Introduction}
\label{introduction}

Question Answering (QA) is a fundamental task in reading comprehension in humans. Comprehending sentences and paragraphs requires a level of abstract understanding that only humans are capable of doing. Asking a question is a way to evaluate a reader on how they understood a given passage and it is also a way to see if a given passage contains information that the question asks. Reading Comprehension (RC) tasks in Natural Language Processing (NLP) is one of the most challenging problems in computer science. RC typically involves making machines understand and comprehend sentences and paragraphs with key features like narration and complex reasoning. The RC tasks also have major problems with synthesizing answers which require background and common-sense knowledge that goes beyond the given paragraphs. There has been significant progress in the past that involves RC with the goal of improving QA tasks. A particular QA dataset called SQuAD~\cite{Rajpurkar2016} (Stanford Question and Answering Dataset) is a popular dataset and - which over the past four years - it has been used to build and improve NLP models. NLP models such as the BERT~\cite{Devlin2018} (Bidirectional Encoder Representation from Transformers) model are widely used in developing and training for the purpose of QA. Our particular focus is to fine-tune and perform hyperparameter benchmarking using pretrained BERT-based models on an RC dataset called DuoRC~\cite{Saha2018}.

With the goal of evaluating BERT-based models on DuoRC, this requires fine-tuning these models with a set of hyperparameters. The BERT-based models in question are the original BERT~\cite{Devlin2018}, ALBERT~\cite{Lan2019}, and LongFormer~\cite{Beltagy2020}. The pre-trained weights of these models are available through the Huggingface library \cite{Wolf2019}. These pretrained weights are then fine-tuned using the DuoRC dataset. Fine-tuning means that we further train the model weights so that it fits into the DuoRC dataset. In Section~\ref{related-work}, we explain the background work in terms of RC and QA and the novel datasets used in the past. We also explain the previous work done of DuoRC and the performance of the models used in this dataset. In Section~\ref{dataset}, the DuoRC dataset is explained in detail and the preprocessing done for the fine-tuning tasks. We then proceed to explain the mechanisms of the language models BERT, ALBERT, and Longformer in Section~\ref{models}. The results are then presented in Section~\ref{results} and we discuss the results in detail on what it entails about the DuoRC dataset. Finally, we conclude in Sections ~\ref{conclusion} and ~\ref{future-work} explaining the outcomes and future work of this research work respectively.

\section{Related Work}
\label{related-work}
There are many QA datasets that have been developed over the past few years and the NLP model development also have made significant progress on improving performance on these datasets. For example, the DuoRC~\cite{Saha2018} dataset attempts to push the challenges of RC. Most QA datasets have short passages that contain descriptive passages rather than passages with temporal reasoning and narration. The SQuAD is a particular example of a QA dataset that has short descriptive passages. Similar datasets are the TriviaQA~\cite{Joshi2017}, HotpotQA~\cite{Yang2018}, and MS MARCO~\cite{Bajaj2016}. Usually, the passages have lexical overlap with the questions which made the current models achieve high reasonable performance. The answers to the questions are typically longer - full sentences - than the DuoRC's answers. Other similar datasets to DuoRC are the MovieQA~\cite{Tapaswi2016}, NarrativeQA~\cite{Kocisky2018}, and NewsQA~\cite{Trischler2016}. There are other datasets with a much more complex structure. The dataset called Discrete Reasoning Over Paragraphs (DROP~\cite{Dua2019}) contains paragraphs, questions, and answers that involve the reader to make discrete operations such as addition, subtraction, comparison, and coreference resolution. There are also recent datasets that focus on temporal and coreferential reasoning which are the TORQUE~\cite{Ning2020} and Quoref~\cite{Dasigi2020}.

Our particular focus is to evaluate BERT, ALBERT, and Longformer on the DuoRC dataset. We want to know how the hyperparameters influence the performance of the models when fine-tuning the pre-trained model weights. Similar studies have been done, e.g., A Robustly Optimized BERT Pretraining Approach which is also known as RoBERTa\cite{Liu2019}. Their studies involved different approaches to pretraining the BERT model architecture to yield the best performance. They indicated that the choice of hyperparameters - similarly here - significantly impacts the performance of the model. The performance was improved by training the model longer and with a larger training batch size. In comparison to our study, we evaluate three BERT-based models on the same dataset rather than evaluating one model on multiple datasets

\section{Dataset}
\label{dataset}

\begin{figure}[htbp!]
 \centering
 \caption{ \label{fig:duorc-data-example} An example of QA pairs of plots with short and long plots. For illustration purposes, the highlighted colored texts (short plot in \textcolor{blue}{blue} and long plot in \textcolor{green}{green}) are the relevant spans of the question and answers shown at the bottom of each plot. Even though the same questions for both plots are the same, some answers (in \textcolor{red}{red}) appears to be inconsistent between the plots. See the DuoRC~\cite{Saha2018} paper for more examples.}
 \includegraphics[width=\textwidth]{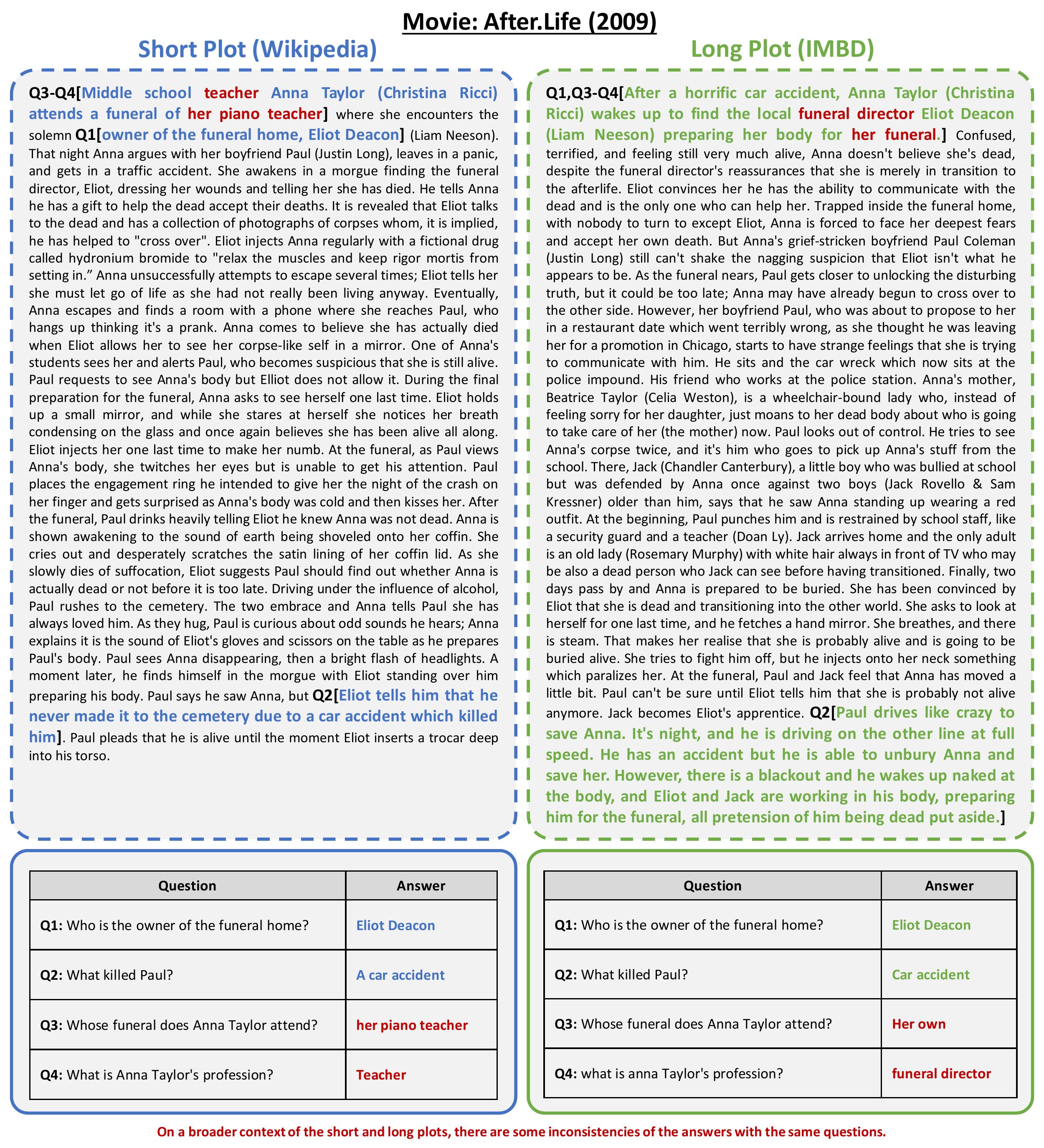}
\end{figure}

\textbf{DuoRC Description.} 
DuoRC~\cite{Saha2018} is a Reading Comprehension (RC) dataset that contains question-answer pairs that are created from pairs of documents containing movie plots. These pairs of documents contain two different versions of the same movie narrative created by different authors. The documents are paired as short and long plots. The short plots are called the original (labeled as ``selfRC'' in DuoRC source code) and the long plot is called ``paraphraseRC''. Both of the documents have the same narrative but with different lengths and word usage. Each pair of documents have the same set of questions and the answers are based on the plot. Fig.~\ref{fig:duorc-data-example} is an illustration of the DuoRC dataset where one plot example is shown with its question and answers. What's shown in this Figure are two versions of the same plot where one is with a short plot taken from Wikipedia and the other is a long plot taken from IMBD. With this Figure, there are four example questions (both plots have the same questions) and answers underneath each plot. For the Short plot, the question Q1 \textit{``Who is the owner of the funeral home?''} has the answer \textit{``Eliot Deacon''} which refers to the context \textit{``...owner of the funeral home, Eliot Deacon...''}. In contrast, the longer plot actually refers to \textit{``Eliot Deacon''} as the \textit{``funeral director''} instead of the \textit{``owner''}. The authors of DuoRC~\cite{Saha2018} collected this dataset where they obtained 7680 movie pairs of long and short plots with 186,089 unique question-answer pairs. The short plots had an average of 580 words while the long plots had an average of 926 words. The QA pairs are created by crowd workers from the Amazon Mechanical Turk (AMT). They first showed the short plot to the first set of workers (2559) to generate QA pairs and they showed the longer plot to a different set of workers (8021) where they answer the questions generated from the first set. The second set of workers also indicated in their answer whether the question is answerable or not based on the context plot. There only 703 workers who are in the first and second sets. The workers are asked to give answers as short as possible.

\textbf{SelfRC vs ParaphraseRC.} 
One of the main differences between the short and long plot is the sentences referred to in the plot for a given question. For example, the question Q2 \textit{``What killed Paul?''} - with answer \textit{``A car accident''} in the short plot and \textit{``Car accident''} in the long plot - refers to two different context sentences from the two plots. The short plot with sentence annotated as \textbf{Q2[...]} (Fig.~\ref{fig:duorc-data-example} left panel) answers the question more directly. In comparison, the long plot with sentence annotated as \textbf{Q2[...]} (Fig.~\ref{fig:duorc-data-example} right panel) has a longer context and did not explicitly say \textit{``Car accident''} but both answers are essentially the same. Our observations and the observations of the authors of DuoRC indicate that there are some inconsistencies and weaknesses in the dataset. Two of them are (1) the plots sometimes have unstructured and inexplicable sentences and (2) some of the answers to the same questions are different between the short and long plot. The different answers are not necessarily a weakness but some have answers which are not correct. On Fig.~\ref{fig:duorc-data-example} (right panel) with \textbf{Q2[...]} annotation, the sentence reads - on some parts - it does not make any sense. Second, the answers to questions Q3-Q4 are different. The short plot answers to Q3-Q4 is consistent with the short plot. However, the answers to the longer plot tell a different outcome. Even though the questions are the same for Q3 \textit{``Whose funeral does Anna Taylor attend''} the answers \textit{``her piano teacher''} versus \textit{``Her own''} refers to two different contexts from the two plots. There is also a wide disagreement on Anna Taylor's profession. The question Q4 \textit{``What is Anna Taylor's profession''} has answer \textit{``Teacher''} in the short plot and \textit{``funeral director''} in the long plot. This is probably a result of human error from the AMT. There is no indication of human performance measures on this particular dataset. In a more broad overview of the data, 40.7\% of the questions have the same answer between the short and long plots, 37.8\% have overlapped, and 21\% have partial overlap. The rest are no-answer questions.

\textbf{Span and Full Subsets.} 
As mentioned in the DuoRC paper, the SelfRC and DuoRC are divided into subsets called ``Span'' and ``Full''. The ``Span'' subset is a set of plots where only the relevant sentences to the questions are extracted. The sentence was considered relevant if at least 50\% of the question words (stop words excluded) overlap with words in the sentence. If there were no 50\% overlap words, the threshold is reduced to 30\% overlap words. The words are considered overlapping if (1) they are verbatim, (2) they are word morphemes, and (3) they are semantically similar, and (4) two words are the same in WordNet. Word morphemes are words that have a root word with pre-fixes, post-fix, or compound words. For example, the words ``unread'', ``readable'', and ``unreadable'' have the same root word ``read''. The semantically similar words are determined using the Glove~\cite{Pennington} and the Skip-thought~\cite{Kiros2015} word embeddings. A word is similar to a word if within the top 50 nearest neighbors of the embedding space. WordNet is a word database - almost similar to a thesaurus - that match a meaningfully similar word. After we additional data processing, the dataset we used for fine-tuning is that we only include questions whose answers are completely verbatim in the context and the un-answerable questions are kept.

\textbf{Previous Models and Reported Performance Results.} The DuoRC paper has its own model that they trained on their dataset. The model in question is the BiDAF~\cite{Seo2016} (BiDirectional Attention Flow) where the model architecture is a multi-stage hierarchical process with bidirectional attention flow. The first BiDAF model is called \textit{SpanModel}. The second BiDAF model is called \textit{GenModel} where the model goes through two stages of processing, span prediction using BiDAF, and answer generation using the span and the question. The performance results of these models on DuoRC are below than the performance on the SQuAD. The BiDAF model was originally trained on the SQuAD dataset. The performance is low due to the fact that the DuoRC dataset has a longer context with a complex structure compared to the SQuAD. We can see in Table~\ref{table:duorc-report} the reported results of the performance of the models trained on DuoRC.

\begin{table}[htbp!]
\centering
\caption{The reported results of the DuoRC~\cite{Saha2018} paper where the performance measures are for the test sets. The values in $[\cdots]$ are reported results in their GitHub repository, \url{https://duorc.github.io/}. The bolded text with a `*' is the measure we can compare with our results in Tables ~\ref{table:selfduorc-results} and ~\ref{table:paraphraseduorc-results}.}
 \label{table:duorc-report}
\begin{tabular}{@{}ccccc@{}}
\toprule
\textbf{DuoRC} &
  \textbf{subset} &
  \textbf{model} &
  \textbf{acc.} &
  \textbf{F1} \\ \midrule
\multicolumn{1}{|c|}{\multirow{4}{*}{\textbf{SelfRC}}} &
  \multicolumn{1}{c|}{\multirow{2}{*}{\textbf{Span}}} &
  \multicolumn{1}{c|}{\textbf{SpanModel}} &
  \multicolumn{1}{c|}{\textbf{46.14*}} &
  \multicolumn{1}{c|}{\textbf{57.49*}} \\ \cmidrule(l){3-5} 
\multicolumn{1}{|c|}{} &
  \multicolumn{1}{c|}{} &
  \multicolumn{1}{c|}{\textbf{GenModel}} &
  \multicolumn{1}{c|}{\textbf{16.45}} &
  \multicolumn{1}{c|}{\textbf{26.97}} \\ \cmidrule(l){2-5} 
\multicolumn{1}{|c|}{} &
  \multicolumn{1}{c|}{\multirow{2}{*}{Full}} &
  \multicolumn{1}{c|}{SpanModel} &
  \multicolumn{1}{c|}{{[}37.53{]}} &
  \multicolumn{1}{c|}{{[}50.56{]}} \\ \cmidrule(l){3-5} 
\multicolumn{1}{|c|}{} &
  \multicolumn{1}{c|}{} &
  \multicolumn{1}{c|}{GenModel} &
  \multicolumn{1}{c|}{{[}15.31{]}} &
  \multicolumn{1}{c|}{{[}24.05{]}} \\ \midrule
\multicolumn{1}{|c|}{\multirow{4}{*}{\textbf{ParaphraseRC}}} &
  \multicolumn{1}{c|}{\multirow{2}{*}{\textbf{Span}}} &
  \multicolumn{1}{c|}{\textbf{SpanModel}} &
  \multicolumn{1}{c|}{\textbf{27.49*}} &
  \multicolumn{1}{c|}{\textbf{35.10*}} \\ \cmidrule(l){3-5} 
\multicolumn{1}{|c|}{} &
  \multicolumn{1}{c|}{} &
  \multicolumn{1}{c|}{\textbf{GenModel}} &
  \multicolumn{1}{c|}{\textbf{12.66}} &
  \multicolumn{1}{c|}{\textbf{19.48}} \\ \cmidrule(l){2-5} 
\multicolumn{1}{|c|}{} &
  \multicolumn{1}{c|}{\multirow{2}{*}{Full}} &
  \multicolumn{1}{c|}{SpanModel} &
  \multicolumn{1}{c|}{{[}14.92{]}} &
  \multicolumn{1}{c|}{{[}21.53{]}} \\ \cmidrule(l){3-5} 
\multicolumn{1}{|c|}{} &
  \multicolumn{1}{c|}{} &
  \multicolumn{1}{c|}{GenModel} &
  \multicolumn{1}{c|}{{[}5.42{]}} &
  \multicolumn{1}{c|}{{[}9.64{]}} \\ \bottomrule
\end{tabular}
\end{table}

This study only uses the Span subset of the SelfRC and ParaphraseRC for fine-tuning the BERT, ALBERT, and LongFormer Models.

\section{Models}
\label{models}

\begin{figure}[htbp!]
 \centering
 \caption{ \label{fig:tokenization-and-hyperparameters} This diagram shows the tokenization process and it shows how the hyperparameter during training works. The hyperparameter labels are in bolded red text.}
 \includegraphics[width=\textwidth]{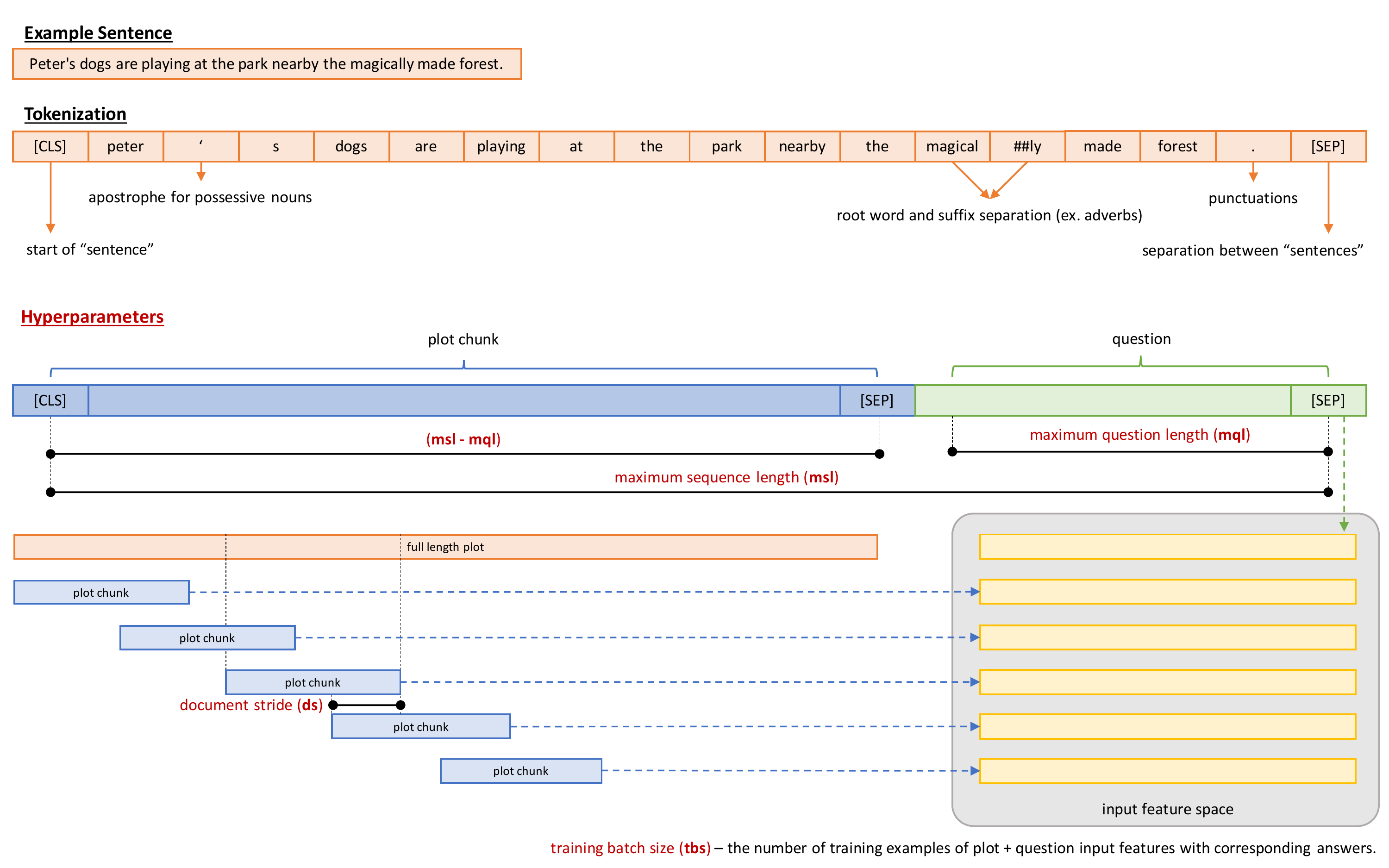}
\end{figure}

\textbf{Tokenization and Hyperparameters.} The tokenization is an important pre-step process for any NLP models. There are many different ways tokenization is done. Specifically for the BERT model, the tokenization is based on word and word structures like the apostrophe for possessive nouns and suffixes of adverbs (see Fig.~\ref{fig:tokenization-and-hyperparameters}). The entire tokenization process is done using WordPiece tokenization \cite{Schuster2012}. This is a subword segmentation algorithm that was originally used for Neural Machine Translation (NMT) tasks and the algorithm is then used in the BERT model \cite{Wu2016}. An additional token ``[CLS']'' is added at the beginning of the sequence and a token ``[SEP]'' which indicated a separation of two ``sentences'' or at the end. A ``sentence'' is applied loosely as an actual sentence or a few paragraphs. The hyperparameters are parameters that we can adjust prior to training. The maximum sequence length (\textbf{msl}) is the maximum length - or number of tokens - of the input on a given model which includes the question and the context. The maximum question length (\textbf{mql}) is the maximum length - or number of tokens - of the question input. If the input plot is longer than the maximum sequence length then the approach is to take chunks of the sequence to the max length with a given document stride. The document stride (\textbf{ds}) is the length of each stride on how the pre-processing of the long context turned into features (see Fig.~\ref{fig:tokenization-and-hyperparameters}). The training batch size (\textbf{tbs}) is the number of training examples of plot and question input features with corresponding answers. It is also the number of batches per GPU. We used 4 GPUs when fine-tuning. There are effectively $4 \times $\textbf{tbs} actual batch size. These hyperparameters are standard for training QA models using the Huggingface library \cite{Wolf2019}.

\textbf{BERT.} The Bidirectional Encoder Representations from Transformers (BERT) \cite{Devlin2018} is a language representation model that uses encoder-decoder architecture. The general structure of these neural networks processes language in sequence and the input passes through a layer called an “encoder” where the words are communicating information with each other for the model to generate a semantic structure. For the model to perform a task such as QA, the flow of information then passes through a “decoder” where it performs the reverse operation while it recovers the information from the “encoder”. This type of neural network structure is known as Transformers in NLP. This is a novel architecture that is designed to process sequential data. Similar architectures such as the recurrent neural networks (or RNNs), typically require it to process the sequence data from beginning to end but Transformers avoids this requirement, and this led to a novel ground-work for faster models to train on large datasets. We refer the reader to the original paper for details. In short, the BERT model has two versions. The first version is $\mathbf{BERT_{BASE}}$ with 12 layers with hidden layer sizes of 768 and 12 attention layers. The second version is $\mathbf{BERT_{LARGE}}$ with 24 layers with hidden layer sizes of 1024 and 16 attention layers. The parameters are approximately 110M and 340M respectively. Self-attention layers use dot-product similarity scores to augment each token’s representation with information from other tokens that are similar or contextually informative. Crucially, self-attention layers are fully parallelizable because there is no recursiveness, unlike RNNs.

\textbf{ALBERT.} The A Lite BERT (ALBERT) \cite{Lan2019} is a smaller modified version of BERT. It uses transformer neural network architecture just like BERT. In ALBERT, the factorization of the embedding parameters is implemented. The model decomposes the embedding space into two lower dimensional matrices before projecting it into the hidden space. In this case, the number of parameters of ALBERT is reduced compared to the original BERT. Similar to BERT, ALBERT has two versions which are ALBERT-large and ALBERT-base. It also has two other versions which are ALBERT-xlarge and ALBERT-xxlarge. The ALBERT-base has approximately 12M parameters with 12 layers where each layer has 768 hidden layer sizes. The ALBERT-large has approximately 18M parameters with 24 layers where each layer has 1024 hidden layer sizes. One key additional feature for ALBERT is the cross-layer parameter sharing. The purpose of the parameter sharing is to improve parameter efficiency. The parameters are shared across layers using a feed-forward network (FFN).

\textbf{LongFormer.} The Long document TransFormer (LongFormer) \cite{Beltagy2020} is designed to deal with long documents. For BERT and ALBERT, the number of token inputs for the model is limited to an upper bound of 512 tokens, the LongFormer model has input tokens upper bound of size 4096. Similar to BERT, the LongFormer has Transformers neural network architecture style with self-attention mechanisms. The key additional feature for LongFormer is the ``Attention Pattern''. Due to the importance of local context, the LongFormer model builds a sliding window approach for attention. The layers are stacked using the windowed attention that scales linearly based on sequence length. A ``global attention'' is inserted into selected input locations so that the model is flexible for training a specific task like QA. Due to this approach, the number of parameters for LongFormer is approximately 149M.

\textbf{Pretrained Weights and Performance Measures.} The pretrained weights for the BERT, ALBERT, and LongFormer models are freely accessible from the Huggingface library \cite{Wolf2019}. The performance measure used in this study is the F1 and accuracy scores. In classification modeling, the F1 score is a measure of accuracy where precision and recall are used to compute it. The precision score is the fraction of true positives over the total of true positive and false positives. The recall score is the fraction of true positive over the total of relevant items. The F1 is the harmonic mean of precision and recall. The accuracy score is correctly classified items over the total.

\section{Results}
\label{results}

\textbf{SelfRC.} We see a significant improvement of the performances across all models for the SelfRC datasets compared to the original performances from Table~\ref{table:duorc-report}. Table~\ref{table:selfduorc-results} shows that the best model is the ALBERT model pretrained with the SQuAD2 dataset. The best F1 score is 76.4 and the accuracy score is 68.52 using the validation set. The results also indicate that increasing the hyperparameters improves the performance of the models. For example, the highest \textbf{msl} hyperparameter we tested yielded the highest performance value across all models. The models with \textbf{tbs} of 4 took longer to fine-tune than the models with \textbf{tbs} of 10. The larger the \textbf{tbs} the better the performance will be. Recall that the \textbf{tbs} is the batch size per GPU. Since we used 4 GPUs during training, it is actually $4 \times$\textbf{tbs} batch size. However, due to memory constraints, we chose a \textbf{tbs} of 4 for the LongFormer models with \textbf{msl} of 768.

\begin{table}[htbp!]
\centering
\caption{\textbf{Short Plot (SelfRC) of DuoRC Fine-Tuning Results.} Each row corresponds to a combination of hyperparameters used to fine-tune the models using the Self DuoRC dataset. The bolded rows indicate the best F1 values for the validation set for each model while the best model overall is marked with a `*'.}
\label{table:selfduorc-results}
\begin{tabular}{@{}rcccccccc@{}}
\toprule
\textbf{model} &
  \textbf{msl} &
  \textbf{ds} &
  \textbf{mql} &
  \textbf{tbs} &
  \textbf{F1 val} &
  \textbf{F1 test} &
  \textbf{acc. val} &
  \textbf{acc. test} \\ \midrule
\multicolumn{1}{|r|}{\multirow{4}{*}{\textbf{\begin{tabular}[c]{@{}r@{}}albert-base-v1\\ (SQuAD1 pretrained)\end{tabular}}}} &
  \multicolumn{1}{c|}{\multirow{2}{*}{384}} &
  \multicolumn{1}{c|}{\multirow{2}{*}{128}} &
  \multicolumn{1}{c|}{\multirow{2}{*}{64}} &
  \multicolumn{1}{c|}{4} &
  \multicolumn{1}{c|}{70.8} &
  \multicolumn{1}{c|}{70.87} &
  \multicolumn{1}{c|}{62.89} &
  \multicolumn{1}{c|}{62.78} \\ \cmidrule(l){5-9} 
\multicolumn{1}{|r|}{} &
  \multicolumn{1}{c|}{} &
  \multicolumn{1}{c|}{} &
  \multicolumn{1}{c|}{} &
  \multicolumn{1}{c|}{10} &
  \multicolumn{1}{c|}{71.13} &
  \multicolumn{1}{c|}{71.03} &
  \multicolumn{1}{c|}{63.28} &
  \multicolumn{1}{c|}{63.2} \\ \cmidrule(l){2-9} 
\multicolumn{1}{|r|}{} &
  \multicolumn{1}{c|}{\multirow{2}{*}{\textbf{512}}} &
  \multicolumn{1}{c|}{128} &
  \multicolumn{1}{c|}{64} &
  \multicolumn{1}{c|}{10} &
  \multicolumn{1}{c|}{74.45} &
  \multicolumn{1}{c|}{74.21} &
  \multicolumn{1}{c|}{66.37} &
  \multicolumn{1}{c|}{66.16} \\ \cmidrule(l){3-9} 
\multicolumn{1}{|r|}{} &
  \multicolumn{1}{c|}{} &
  \multicolumn{1}{c|}{\textbf{384}} &
  \multicolumn{1}{c|}{\textbf{64}} &
  \multicolumn{1}{c|}{\textbf{10}} &
  \multicolumn{1}{c|}{\textbf{74.47}} &
  \multicolumn{1}{c|}{\textbf{74.33}} &
  \multicolumn{1}{c|}{\textbf{66.17}} &
  \multicolumn{1}{c|}{\textbf{66.23}} \\ \midrule
\multicolumn{1}{|r|}{\multirow{4}{*}{\textbf{\begin{tabular}[c]{@{}r@{}}albert-base-v2*\\ (SQuAD2 pretrained)\end{tabular}}}} &
  \multicolumn{1}{c|}{\multirow{2}{*}{384}} &
  \multicolumn{1}{c|}{\multirow{2}{*}{128}} &
  \multicolumn{1}{c|}{\multirow{2}{*}{64}} &
  \multicolumn{1}{c|}{4} &
  \multicolumn{1}{c|}{68.35} &
  \multicolumn{1}{c|}{69.46} &
  \multicolumn{1}{c|}{60.38} &
  \multicolumn{1}{c|}{61.78} \\ \cmidrule(l){5-9} 
\multicolumn{1}{|r|}{} &
  \multicolumn{1}{c|}{} &
  \multicolumn{1}{c|}{} &
  \multicolumn{1}{c|}{} &
  \multicolumn{1}{c|}{10} &
  \multicolumn{1}{c|}{70.91} &
  \multicolumn{1}{c|}{71.56} &
  \multicolumn{1}{c|}{62.67} &
  \multicolumn{1}{c|}{63.81} \\ \cmidrule(l){2-9} 
\multicolumn{1}{|r|}{} &
  \multicolumn{1}{c|}{\multirow{2}{*}{\textbf{512*}}} &
  \multicolumn{1}{c|}{128} &
  \multicolumn{1}{c|}{64} &
  \multicolumn{1}{c|}{10} &
  \multicolumn{1}{c|}{75.33} &
  \multicolumn{1}{c|}{75.07} &
  \multicolumn{1}{c|}{67.48} &
  \multicolumn{1}{c|}{67.22} \\ \cmidrule(l){3-9} 
\multicolumn{1}{|r|}{} &
  \multicolumn{1}{c|}{} &
  \multicolumn{1}{c|}{\textbf{384*}} &
  \multicolumn{1}{c|}{\textbf{64*}} &
  \multicolumn{1}{c|}{\textbf{10*}} &
  \multicolumn{1}{c|}{\textbf{76.4*}} &
  \multicolumn{1}{c|}{\textbf{76.29*}} &
  \multicolumn{1}{c|}{\textbf{68.52*}} &
  \multicolumn{1}{c|}{\textbf{68.32*}} \\ \midrule
\multicolumn{1}{|r|}{\multirow{4}{*}{\textbf{\begin{tabular}[c]{@{}r@{}}bert-base-uncased\\ (BooksCorpus and\\  English Wiki pretrained)\end{tabular}}}} &
  \multicolumn{1}{c|}{\multirow{2}{*}{384}} &
  \multicolumn{1}{c|}{\multirow{2}{*}{128}} &
  \multicolumn{1}{c|}{\multirow{2}{*}{64}} &
  \multicolumn{1}{c|}{4} &
  \multicolumn{1}{c|}{70.69} &
  \multicolumn{1}{c|}{70.2} &
  \multicolumn{1}{c|}{62.5} &
  \multicolumn{1}{c|}{62.01} \\ \cmidrule(l){5-9} 
\multicolumn{1}{|r|}{} &
  \multicolumn{1}{c|}{} &
  \multicolumn{1}{c|}{} &
  \multicolumn{1}{c|}{} &
  \multicolumn{1}{c|}{10} &
  \multicolumn{1}{c|}{70.56} &
  \multicolumn{1}{c|}{70.62} &
  \multicolumn{1}{c|}{62.26} &
  \multicolumn{1}{c|}{62.35} \\ \cmidrule(l){2-9} 
\multicolumn{1}{|r|}{} &
  \multicolumn{1}{c|}{\multirow{2}{*}{\textbf{512}}} &
  \multicolumn{1}{c|}{\textbf{128}} &
  \multicolumn{1}{c|}{\textbf{64}} &
  \multicolumn{1}{c|}{\textbf{10}} &
  \multicolumn{1}{c|}{\textbf{74.14}} &
  \multicolumn{1}{c|}{\textbf{73.72}} &
  \multicolumn{1}{c|}{\textbf{65.82}} &
  \multicolumn{1}{c|}{\textbf{65.25}} \\ \cmidrule(l){3-9} 
\multicolumn{1}{|r|}{} &
  \multicolumn{1}{c|}{} &
  \multicolumn{1}{c|}{384} &
  \multicolumn{1}{c|}{64} &
  \multicolumn{1}{c|}{10} &
  \multicolumn{1}{c|}{73.63} &
  \multicolumn{1}{c|}{73.07} &
  \multicolumn{1}{c|}{65.06} &
  \multicolumn{1}{c|}{64.72} \\ \midrule
\multicolumn{1}{|r|}{\multirow{6}{*}{\textbf{\begin{tabular}[c]{@{}r@{}}longformer-squad1\\ (SQuAD1 pretrained)\end{tabular}}}} &
  \multicolumn{1}{c|}{\multirow{2}{*}{384}} &
  \multicolumn{1}{c|}{128} &
  \multicolumn{1}{c|}{64} &
  \multicolumn{1}{c|}{4} &
  \multicolumn{1}{c|}{67.99} &
  \multicolumn{1}{c|}{67.83} &
  \multicolumn{1}{c|}{59.82} &
  \multicolumn{1}{c|}{59.36} \\ \cmidrule(l){3-9} 
\multicolumn{1}{|r|}{} &
  \multicolumn{1}{c|}{} &
  \multicolumn{1}{c|}{128} &
  \multicolumn{1}{c|}{64} &
  \multicolumn{1}{c|}{10} &
  \multicolumn{1}{c|}{67.47} &
  \multicolumn{1}{c|}{68.16} &
  \multicolumn{1}{c|}{59.38} &
  \multicolumn{1}{c|}{59.77} \\ \cmidrule(l){2-9} 
\multicolumn{1}{|r|}{} &
  \multicolumn{1}{c|}{\multirow{2}{*}{512}} &
  \multicolumn{1}{c|}{128} &
  \multicolumn{1}{c|}{64} &
  \multicolumn{1}{c|}{10} &
  \multicolumn{1}{c|}{71.07} &
  \multicolumn{1}{c|}{70.19} &
  \multicolumn{1}{c|}{62.72} &
  \multicolumn{1}{c|}{61.76} \\ \cmidrule(l){3-9} 
\multicolumn{1}{|r|}{} &
  \multicolumn{1}{c|}{} &
  \multicolumn{1}{c|}{384} &
  \multicolumn{1}{c|}{64} &
  \multicolumn{1}{c|}{10} &
  \multicolumn{1}{c|}{70.92} &
  \multicolumn{1}{c|}{70.59} &
  \multicolumn{1}{c|}{62.59} &
  \multicolumn{1}{c|}{62.21} \\ \cmidrule(l){2-9} 
\multicolumn{1}{|r|}{} &
  \multicolumn{1}{c|}{\multirow{2}{*}{\textbf{768}}} &
  \multicolumn{1}{c|}{128} &
  \multicolumn{1}{c|}{64} &
  \multicolumn{1}{c|}{4} &
  \multicolumn{1}{c|}{73.47} &
  \multicolumn{1}{c|}{73.39} &
  \multicolumn{1}{c|}{65.08} &
  \multicolumn{1}{c|}{64.97} \\ \cmidrule(l){3-9} 
\multicolumn{1}{|r|}{} &
  \multicolumn{1}{c|}{} &
  \multicolumn{1}{c|}{\textbf{384}} &
  \multicolumn{1}{c|}{\textbf{64}} &
  \multicolumn{1}{c|}{\textbf{4}} &
  \multicolumn{1}{c|}{\textbf{75.9}} &
  \multicolumn{1}{c|}{\textbf{75.06}} &
  \multicolumn{1}{c|}{\textbf{67.24}} &
  \multicolumn{1}{c|}{\textbf{66.41}} \\ \midrule
\multicolumn{1}{|r|}{\multirow{6}{*}{\textbf{\begin{tabular}[c]{@{}r@{}}longformer-squad2\\ (SQuAD2 pretrained)\end{tabular}}}} &
  \multicolumn{1}{c|}{\multirow{2}{*}{384}} &
  \multicolumn{1}{c|}{\multirow{2}{*}{128}} &
  \multicolumn{1}{c|}{\multirow{2}{*}{64}} &
  \multicolumn{1}{c|}{4} &
  \multicolumn{1}{c|}{67.92} &
  \multicolumn{1}{c|}{67.56} &
  \multicolumn{1}{c|}{59.33} &
  \multicolumn{1}{c|}{59.02} \\ \cmidrule(l){5-9} 
\multicolumn{1}{|r|}{} &
  \multicolumn{1}{c|}{} &
  \multicolumn{1}{c|}{} &
  \multicolumn{1}{c|}{} &
  \multicolumn{1}{c|}{10} &
  \multicolumn{1}{c|}{69.06} &
  \multicolumn{1}{c|}{69.08} &
  \multicolumn{1}{c|}{60.71} &
  \multicolumn{1}{c|}{60.83} \\ \cmidrule(l){2-9} 
\multicolumn{1}{|r|}{} &
  \multicolumn{1}{c|}{\multirow{2}{*}{512}} &
  \multicolumn{1}{c|}{128} &
  \multicolumn{1}{c|}{64} &
  \multicolumn{1}{c|}{10} &
  \multicolumn{1}{c|}{71.82} &
  \multicolumn{1}{c|}{71.33} &
  \multicolumn{1}{c|}{63.21} &
  \multicolumn{1}{c|}{62.66} \\ \cmidrule(l){3-9} 
\multicolumn{1}{|r|}{} &
  \multicolumn{1}{c|}{} &
  \multicolumn{1}{c|}{384} &
  \multicolumn{1}{c|}{64} &
  \multicolumn{1}{c|}{10} &
  \multicolumn{1}{c|}{70.86} &
  \multicolumn{1}{c|}{69.86} &
  \multicolumn{1}{c|}{62.4} &
  \multicolumn{1}{c|}{61.53} \\ \cmidrule(l){2-9} 
\multicolumn{1}{|r|}{} &
  \multicolumn{1}{c|}{\multirow{2}{*}{\textbf{768}}} &
  \multicolumn{1}{c|}{\textbf{128}} &
  \multicolumn{1}{c|}{\textbf{64}} &
  \multicolumn{1}{c|}{\textbf{4}} &
  \multicolumn{1}{c|}{\textbf{74.35}} &
  \multicolumn{1}{c|}{\textbf{73.64}} &
  \multicolumn{1}{c|}{\textbf{65.7}} &
  \multicolumn{1}{c|}{\textbf{65.22}} \\ \cmidrule(l){3-9} 
\multicolumn{1}{|r|}{} &
  \multicolumn{1}{c|}{} &
  \multicolumn{1}{c|}{384} &
  \multicolumn{1}{c|}{64} &
  \multicolumn{1}{c|}{4} &
  \multicolumn{1}{c|}{73.92} &
  \multicolumn{1}{c|}{73.82} &
  \multicolumn{1}{c|}{65.34} &
  \multicolumn{1}{c|}{65.15} \\ \bottomrule
\end{tabular}
\end{table}

\textbf{ParaphraseRC.} Table~\ref{table:paraphraseduorc-results} shows that the best model is the LongFormer model pretrained with the SelfRC dataset. The best F1 score is 52.78 and the accuracy score is 46.60 using the validation set. The hyperparameters used in fine-tuning the ParaphraseRC dataset is chosen from the results of the SelfRC dataset fine-tuning. The F1 scores for the Paraphrase RC are lower than the results for the SelfRC as expected. Compared to the selfRC, the ParaphraseRC dataset contains longer plots with questions that may or may not have an overlapping vocabulary.

\begin{table}[htbp!]
\centering
\caption{\textbf{Long Plot (ParaphraseRC) DuoRC Fine-Tuning Results.} Each row corresponds to a combination of hyperparameters used to fine-tune the models using the Paraphrase DuoRC dataset. The bolded rows and marked with `*' indicates the best F1 value for the validation set.}
\label{table:paraphraseduorc-results}
\begin{tabular}{@{}rcccccccc@{}}
\toprule
\textbf{model} &
  \textbf{msl} &
  \textbf{ds} &
  \textbf{mql} &
  \textbf{tbs} &
  \textbf{F1 val} &
  \textbf{F1 test} &
  \textbf{acc. val} &
  \textbf{acc. test} \\ \midrule
\multicolumn{1}{|r|}{\begin{tabular}[c]{@{}r@{}}albert-base-v1\\ (SQuAD1 pretrained)\end{tabular}} &
  \multicolumn{1}{c|}{512} &
  \multicolumn{1}{c|}{384} &
  \multicolumn{1}{c|}{64} &
  \multicolumn{1}{c|}{10} &
  \multicolumn{1}{c|}{50.31} &
  \multicolumn{1}{c|}{51.08} &
  \multicolumn{1}{c|}{45.15} &
  \multicolumn{1}{c|}{45.86} \\ \midrule
\multicolumn{1}{|r|}{\begin{tabular}[c]{@{}r@{}}albert-base-v2\\ (SQuAD3 pretrained)\end{tabular}} &
  \multicolumn{1}{c|}{512} &
  \multicolumn{1}{c|}{384} &
  \multicolumn{1}{c|}{64} &
  \multicolumn{1}{c|}{10} &
  \multicolumn{1}{c|}{50.14} &
  \multicolumn{1}{c|}{50.91} &
  \multicolumn{1}{c|}{45.15} &
  \multicolumn{1}{c|}{45.6} \\ \midrule
\multicolumn{1}{|r|}{\begin{tabular}[c]{@{}r@{}}albert-selfduorc-v1\\ (Self DuoRC pretrained)\end{tabular}} &
  \multicolumn{1}{c|}{512} &
  \multicolumn{1}{c|}{384} &
  \multicolumn{1}{c|}{64} &
  \multicolumn{1}{c|}{10} &
  \multicolumn{1}{c|}{51.25} &
  \multicolumn{1}{c|}{51.74} &
  \multicolumn{1}{c|}{45.94} &
  \multicolumn{1}{c|}{46.31} \\ \midrule
\multicolumn{1}{|r|}{\begin{tabular}[c]{@{}r@{}}albert-selfduorc-v2\\ (Self DuoRC pretrained)\end{tabular}} &
  \multicolumn{1}{c|}{512} &
  \multicolumn{1}{c|}{384} &
  \multicolumn{1}{c|}{64} &
  \multicolumn{1}{c|}{10} &
  \multicolumn{1}{c|}{51.41} &
  \multicolumn{1}{c|}{51.92} &
  \multicolumn{1}{c|}{46.34} &
  \multicolumn{1}{c|}{46.48} \\ \midrule
\multicolumn{1}{|r|}{\begin{tabular}[c]{@{}r@{}}bert-base-uncased\\ (BooksCorpus and \\ English Wiki pretrained)\end{tabular}} &
  \multicolumn{1}{c|}{512} &
  \multicolumn{1}{c|}{128} &
  \multicolumn{1}{c|}{64} &
  \multicolumn{1}{c|}{10} &
  \multicolumn{1}{c|}{48.8} &
  \multicolumn{1}{c|}{48.22} &
  \multicolumn{1}{c|}{43.1} &
  \multicolumn{1}{c|}{42.63} \\ \midrule
\multicolumn{1}{|r|}{\begin{tabular}[c]{@{}r@{}}bert-selfduorc-uncased\\ (Self DuoRC pretrained)\end{tabular}} &
  \multicolumn{1}{c|}{512} &
  \multicolumn{1}{c|}{128} &
  \multicolumn{1}{c|}{64} &
  \multicolumn{1}{c|}{10} &
  \multicolumn{1}{c|}{51.24} &
  \multicolumn{1}{c|}{51.09} &
  \multicolumn{1}{c|}{45.37} &
  \multicolumn{1}{c|}{45.2} \\ \midrule
\multicolumn{1}{|r|}{\begin{tabular}[c]{@{}r@{}}longformer-squad1\\ (SQuAD1 pretrained)\end{tabular}} &
  \multicolumn{1}{c|}{768} &
  \multicolumn{1}{c|}{384} &
  \multicolumn{1}{c|}{64} &
  \multicolumn{1}{c|}{4} &
  \multicolumn{1}{c|}{50.68} &
  \multicolumn{1}{c|}{50.07} &
  \multicolumn{1}{c|}{44.06} &
  \multicolumn{1}{c|}{43.55} \\ \midrule
\multicolumn{1}{|r|}{\begin{tabular}[c]{@{}r@{}}longformer-squad2\\ (SQuAD2 pretrained)\end{tabular}} &
  \multicolumn{1}{c|}{768} &
  \multicolumn{1}{c|}{384} &
  \multicolumn{1}{c|}{64} &
  \multicolumn{1}{c|}{4} &
  \multicolumn{1}{c|}{51.68} &
  \multicolumn{1}{c|}{50.69} &
  \multicolumn{1}{c|}{45.26} &
  \multicolumn{1}{c|}{44.17} \\ \midrule
\multicolumn{1}{|r|}{\textbf{\begin{tabular}[c]{@{}r@{}}longformer-selfduorc1*\\ (Self DuoRC pretrained)\end{tabular}}} &
  \multicolumn{1}{c|}{\textbf{768*}} &
  \multicolumn{1}{c|}{\textbf{384*}} &
  \multicolumn{1}{c|}{\textbf{64*}} &
  \multicolumn{1}{c|}{\textbf{4*}} &
  \multicolumn{1}{c|}{\textbf{52.78*}} &
  \multicolumn{1}{c|}{\textbf{51.94*}} &
  \multicolumn{1}{c|}{\textbf{46.60*}} &
  \multicolumn{1}{c|}{\textbf{45.22*}} \\ \midrule
\multicolumn{1}{|r|}{\begin{tabular}[c]{@{}r@{}}longformer-selfduorc2\\ (Self DuoRC pretrained)\end{tabular}} &
  \multicolumn{1}{c|}{768} &
  \multicolumn{1}{c|}{384} &
  \multicolumn{1}{c|}{64} &
  \multicolumn{1}{c|}{4} &
  \multicolumn{1}{c|}{52.28} &
  \multicolumn{1}{c|}{52} &
  \multicolumn{1}{c|}{45.82} &
  \multicolumn{1}{c|}{45.52} \\ \bottomrule
\end{tabular}
\end{table}

\section{Conclusion}
\label{conclusion}

We have performed a grid search hyperparameter benchmarking on three models on the DuoRC dataset. The models we evaluated are the BERT, ALBERT, and LongFormer models which are transformer-based neural network models. The DuoRC dataset contained two main components for each unique plot. The SelfRC has shorter plot lines while the Paraphrase has longer plot lines. The dataset was reduced into subsets called the ``span'' and ``Full'' where the ``span'' subset is the set of plots where only the relevant sentences to the questions are extracted. The best performing model is the ALBERT model which was pretrained using the SQuAD1 and fine-tuned on the SelfRC. The best performing model fine-tuned on the ParaphraseRC is the LongFormer model which was pretrained using the SelfRC.

\section{Future Work}
\label{future-work}

Our work is only a step toward improving NLP models for QA and RC in general. Increasing the \textbf{msl} hyperparameter for the LongFormer model would definitely improve the performance. The LongFormer model is specifically designed for datasets with longer context documents and should be considered on model development for RC. Datasets like DuoRC contain mostly words with narrative structures but other datasets that contain scientific vocabulary, mathematical equations, logic may be problematic when fine-tuning models for RC. Pretrained models may not always work on a different dataset which is why fine-tuning is the key to improve models to do a specific task.

\section*{Acknowledgments}
This work was performed under the auspices of the U.S. Department of Energy by Lawrence Livermore National Laboratory (LLNL) under contract DE AC52 07NA27344. The first author was supported as a summer intern at LLNL. LLNL-TR-817729.

\bibliographystyle{siam}
\bibliography{references} 

%
%
%
%

\end{document}